\documentclass[11pt]{article}
\usepackage{fullpage}

\usepackage{booktabs}
\usepackage{color}

\usepackage{algorithm}
\usepackage{algorithmic}

\usepackage{url}

\usepackage{amssymb}
\usepackage{amsmath}
\usepackage{mathtools}
\usepackage{multicol}
\usepackage{multirow}

\newcommand{\LabelFunc}[2]{\parbox[t]{#2\textwidth}%
	{\hspace*{\fill}{\vspace*{-0.3cm}#1}\hspace*{\fill}}}

\newcommand{\ThreeLabels}[3]{\LabelFunc{#1}{0.31}\hfill
                             \LabelFunc{#2}{0.31}\hfill\LabelFunc{#3}{0.31}}

\newcommand{\Stiefel}[2]{{\mathrm{St}({#1},{#2})}}

\newcommand{\hess}{\mathrm{Hess}}

\newcommand{\inner}[1]{\left\langle#1\right\rangle}

\def\E{\mathcal{E}}
\def\G{\mathcal{G}}

\def\BB{\mathbb{B}}
\def\HH{\mathbb{H}}
\def\R{\mathbb{R}}

\def\L{\mathcal{L}}

\newcommand{\norm}[1]{\left\|#1\right\|}

\def\grad{\mathop{\rm grad}\nolimits}

\def\bzero{{\mathbf 0}}

\def\bu{\mathbf u}
\def\bv{\mathbf v}
\def\bw{{\mathbf w}}
\def\bx{{\mathbf x}}

\def\bz{{\mathbf z}}

\def\bA{\mathbf A}

\def\bI{{\mathbf I}}

\def\bL{\mathbf L}

\def\bU{{\mathbf U}}

\def\bX{{\mathbf X}}

\def\bZ{{\mathbf Z}}

\def\Poincare{Poincar{\'e} }

\def\minop{\mathop{\rm min}\limits}

\def\min{\mathop{\rm min}\nolimits}

\begin{document}
\title{\LARGE \bf
Low-rank approximations of hyperbolic embeddings\footnote{The authors are with Microsoft, India. Emails: $\{$prjawanp, mamegh, bamdevm$\}$@microsoft.com.}
}


\author{Pratik Jawanpuria \and Mayank Meghwanshi  \and Bamdev Mishra}


\maketitle

\begin{abstract}
The hyperbolic manifold is a smooth manifold of negative constant curvature. While the hyperbolic manifold is well-studied in the literature, it has gained interest in the machine learning and natural language processing communities lately due to its usefulness in modeling continuous hierarchies. Tasks with hierarchical structures are ubiquitous in those fields and there is a general interest to learning hyperbolic representations or embeddings of such tasks. Additionally, these embeddings of related tasks may also share a low-rank subspace. In this work, we propose to learn hyperbolic embeddings such that they also lie in a low-dimensional subspace. In particular, we consider the problem of learning a low-rank factorization of hyperbolic embeddings. We cast these problems as manifold optimization problems and propose computationally efficient algorithms. Empirical results illustrate the efficacy of the proposed approach.
\end{abstract}

\section{Introduction}
Learning hyperbolic representation of entities have gained recent interest in the machine learning community \cite{bronstein17a,muscoloni17a,nickel17a,sala18a,gu19a}. In particular, hyperbolic embeddings have been shown to be well-suited for various natural language processing problems \cite{nickel17a,dhingra18a,ganea19a} that require modeling hierarchical structures such as knowledge graphs, hypernymy hierarchies, organization hierarchy, and taxonomies, among others. The reason being learning representations in the hyperbolic space provides a principled approach for integrating structural information encoded in such (discrete) entities into continuous space. 

\begin{figure*}[t]\centering
{
\includegraphics[width=0.33\columnwidth]{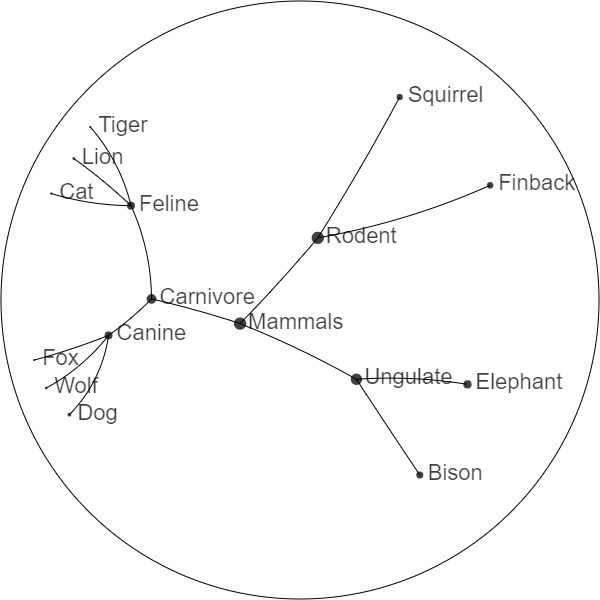}%
\hspace*{\fill}
\includegraphics[width=0.33\columnwidth]{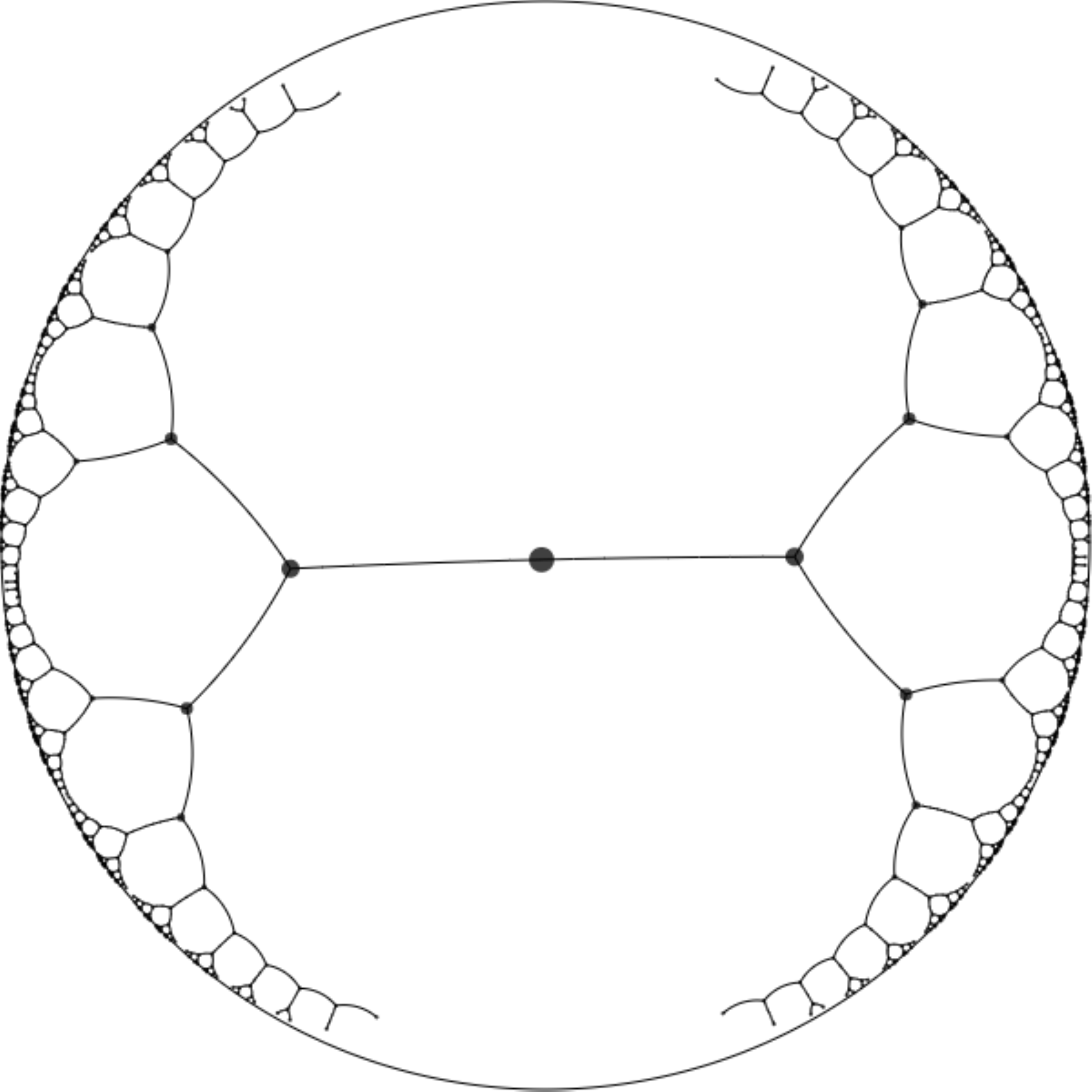}%
\hspace*{\fill}
\includegraphics[width=0.34\columnwidth]{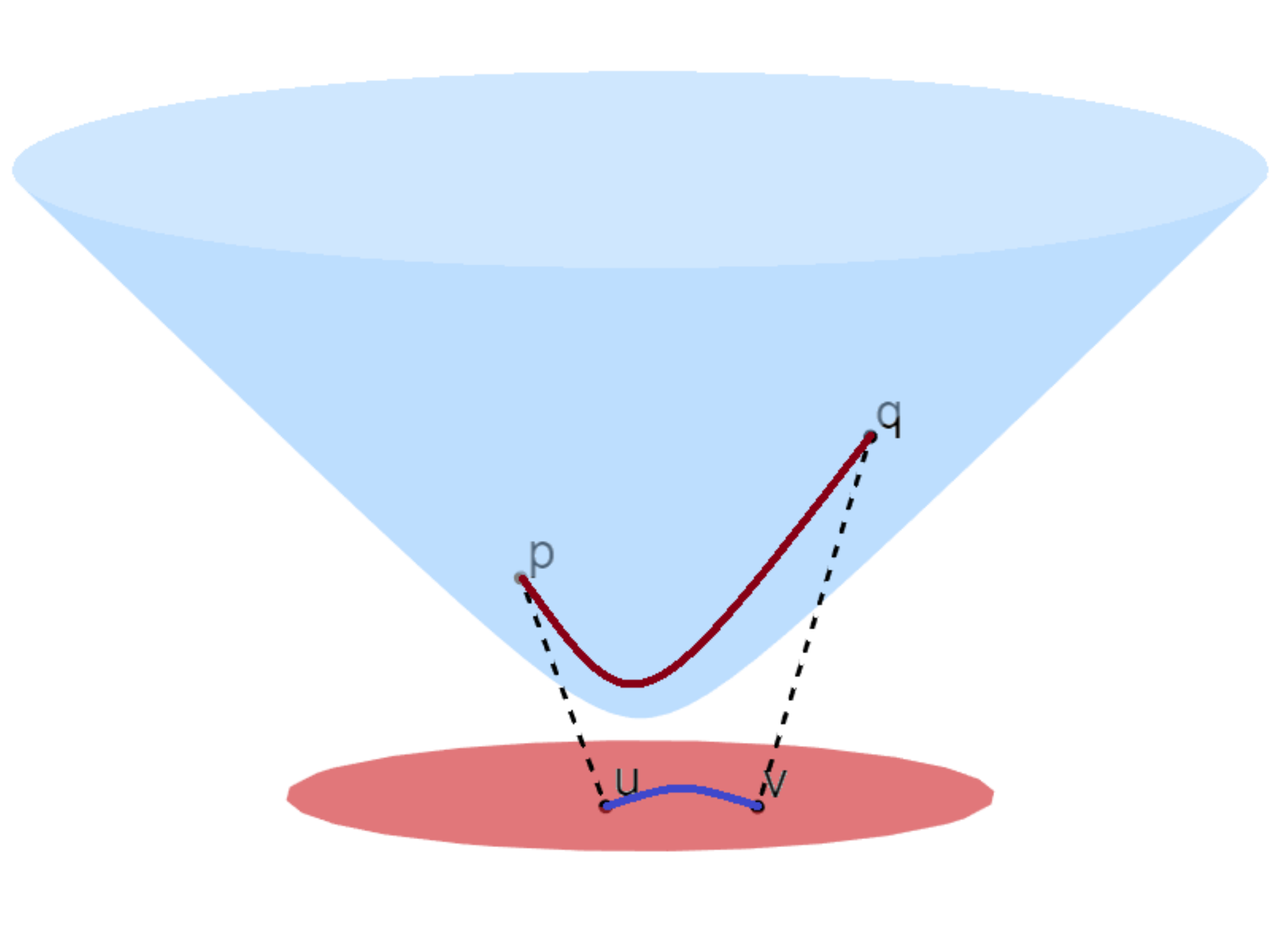}%
}
{\ThreeLabels {(a)} {(b)} {(c)}}
\caption{(a) An example of the hyperbolic space ($2$-dimensional \Poincare ball model $\BB^2$) being used to represent a mammal taxonomy. This taxonomy is a part of WordNet \cite{miller95}; (b) A tree is embedded in $\BB^2$. The two subtrees from the root are regular trees. All the edges have the same hyperbolic length, computed using (\ref{eqn:poincare-distance}); (c) The \Poincare disk ($\BB^2$) may be viewed as a stereoscopic projection of the hyperboloid model ($\HH^2$). Points $p$ and $q$ lie on $\HH^2$ and points $u$ and $v$ are their projections, respectively, onto the $\BB^2$. The maroon curve $\HH^2$ is the geodesic between $p$ and $q$, which projects to the blue geodesic path between $u$ and $v$ on $\BB^2$. Figure best viewed in color.}\label{fig:hyperbolicDiagrams}
\end{figure*}

The hyperbolic space is a non-Euclidean space and has constant negative curvature. The latter property enables it to \textit{grow} exponentially even in dimension as low as two. Hence, the hyperbolic space has been considered to model trees and complex networks, among others \cite{krioukov10a,hamann18a}. Figure~\ref{fig:hyperbolicDiagrams}(a) is an example of representing a part of mammal taxonomy tree in a hyperbolic space (two-dimensional \Poincare ball). Hyperbolic embeddings (numerical representations of tasks) have been considered in several applications such as question answering system \cite{tay18a}, recommender systems \cite{vinh18a,chamberlain19a}, link prediction \cite{nickel18a,ganea18a}, natural language inference \cite{ganea18b}, vertex classification \cite{chamberlain17a}, and machine translation \cite{gulcehre19a}. 

In this paper, we consider the setting in which an additional low-rank structure may also exist among the learned hyperbolic embeddings. Such a setting may arise when the hierarchical entities are closely related. We propose to learn a low-rank approximation of the given (high dimensional) hyperbolic embeddings. Conceptually, we model high dimensional hyperbolic embeddings as a product of a low-dimensional subspace and low-dimensional hyperbolic embeddings. The optimization problem is cast on the product of the Stiefel and hyperbolic manifolds. We develop an efficient Riemannian trust-region algorithm for solving it. We evaluate the proposed approach on real-world datasets: on the problem of reconstructing taxonomy hierarchies from the embeddings. We observe that the performance of proposed approach match the original embeddings even in low-rank settings. 

The outline of the paper is as follows. Section~\ref{sec:background} discusses two popular models of representing hyperbolic space in the Euclidean setting. In Section~\ref{sec:low-rank}, we present our formulation to approximate given hyperbolic embeddings in a low-rank setting. The optimization algorithm is discussed in Section~\ref{sec:optimization}. The experimental results are presented in Section~\ref{sec:experiment} and Section~\ref{sec:conclusion} concludes the paper.



\section{Background}\label{sec:background}
In this section, we briefly discuss the basic concepts of hyperbolic geometry. Interested readers may refer \cite{anderson05,ratcliffe06} for more details. 

The hyperbolic space (of dimension $\geq2$) is a Riemannian manifold with a constant negative sectional curvature. Similar to the Euclidean or spherical spaces, it is isotropic. However, the Euclidean space is flat (zero curvature) and the spherical space is positively curved. As a result of negative curvature, the circumference and area of a circle in hyperbolic space grow exponentially with the radius. In contrast, the circumference and area of a circle in the hyperbolic space grow linearly and quadratically, respectively, in Euclidean setting. Hence, hyperbolic spaces \textit{expand} faster than the Euclidean spaces. Informally, hyperbolic spaces may be viewed as a continuous counterpart to discrete trees as the metric properties of a two-dimensional hyperbolic space and a $b$-ary tree (a tree with branching factor $b$) are similar. Hence, trees can be embedded into a two-dimensional hyperbolic space while keeping the overall distortion arbitrarily small. In contrast, Euclidean spaces cannot attain this result even with unbounded number of dimensions. 

Since hyperbolic models cannot be represented within Euclidean space without distortion, several (equivalent) models exist for representing hyperbolic spaces for computation purpose. The models are conformal to the Euclidean space and points in one model can be transformed to be represented in another model, while preserving geometric properties such as isometry. However, no model captures all the properties of the hyperbolic geometry. Two hyperbolic models, in particular, have received much interest recently in the machine learning community: the \Poincare  ball model and the hyperboloid model. 

\subsection{\Poincare  ball model}
The \Poincare  ball is a $n$-dimensional hyperbolic space defined as the interior of the $n$-dimensional unit (Euclidean) ball: 
\begin{equation*}
\BB^n=\{\bu\in\R^n|\norm{\bu}<1\},
\end{equation*}
where $\norm{\cdot}$ denotes the Euclidean norm. The distance between two points $\bu,\bv\in\BB^n$ in the \Poincare  ball model is given by 
\begin{equation}\label{eqn:poincare-distance}
d_{\BB}(\bu,\bv)=\mathrm{arccosh}\left(1+2\frac{\norm{\bu-\bv}^2}{(1-\norm{\bu}^2)(1-\norm{\bv}^2)}\right)
\end{equation}
and the \Poincare norm is given by 
\begin{equation*}
\norm{\bu}_\BB\coloneqq d_{\BB}(\bzero,\bu)=2\,\mathrm{arctanh}(\norm{\bu}).
\end{equation*}
We observe that the distance between a pair of points near the boundary of the \Poincare ball (Euclidean norm close to unity) grows much faster than distance between the points close to the center (Euclidean norm close to zero). 
In addition, the distance within the \Poincare ball varies smoothly with respect to points $\bu$ and $\bv$. These properties are helpful embedding discrete  hierarchical structures such as trees in hyperbolic spaces and obtain continuous embeddings which respect the tree metric structure. For instance, the origin of the \Poincare  ball may be mapped to the root node of the tree as the root node is relatively closer to all other nodes (points). The leaf nodes can be places near the boundary to ensure they are relatively distant from other leaf nodes. Additionally, the shortest path between a pair of points is usually via a point closer to the origin, just as the shortest path between two nodes in a tree is via their parent nodes. Figure~\ref{fig:hyperbolicDiagrams}(b) shows a \Poincare disk ($\BB^2$) embedding a tree with two regular subtrees. 


\subsection{Hyperboloid model}
Let $\bar{\bu},\bar{\bv}\in\R^{n+1}$ such that 
$\bar{\bu}=\begin{bmatrix}
u_0\\\bu
\end{bmatrix}$ and 
$\bar{\bv}=\begin{bmatrix}
v_0\\\bv
\end{bmatrix}$ and $\bu,\bv\in\R^n$. 
The Lorentz scalar product $\inner{\cdot,\cdot}_\L$ of two vectors $\bar{\bu}$ and $\bar{\bv}$ is defined as 
\begin{equation}\label{eqn:lorentz_metric}
\inner{\bar{\bu},\bar{\bv}}_\L=\bar{\bu}^\top \bL\bar{\bv}=-u_0v_0 + \bu^\top\bv,
\end{equation} 
where $\bL$ is a $(n+1)$-dimensional diagonal matrix 
\begin{equation*}
\bL= \begin{bmatrix} -1 & \bzero_n^\top \\ \bzero_n & \bI_n \end{bmatrix},
\end{equation*}
$\bzero_n$ is the $n$-dimensional zero column vector, and $\bI_n$ is the $n$-dimensional identity matrix. 

The hyperboloid model, also known as the Lorentz model of hyperbolic geometry, is given by 
\begin{equation*}
\HH^n=\{\bar{\bu}\in\R^{n+1}|\inner{\bar{\bu},\bar{\bu}}_{\L}=-1,u_0>0\}.
\end{equation*}
The model represents the upper sheet of an $n$-dimensional hyperboloid. From the constraint set, it can be observed that if $\bar{\bu}\in\HH^n$, then $u_0=\sqrt{1+\bu^\top\bu}$.

The distance between two points $\bar{\bu},\bar{\bv}\in\HH^n$ in the hyperboloid model is given by 
\begin{equation}\label{eqn:distance_hyperbolic}
d_{\HH}(\bar{\bu},\bar{\bv})=\mathrm{arccosh}(-\inner{\bar{\bu},\bar{\bv}}_\L).
\end{equation}

As stated earlier, both the \Poincare  ball and the hyperboloid models are equivalent and a mapping exists from one model to another \cite{wilson18a}. 
Points on the hyperboloid can be mapped to the \Poincare  ball by 
\begin{equation*}
h:\HH^n\rightarrow\BB^n, h(\bar{\bu})=\frac{\bu}{u_0+1}, \bar{\bu}\in\HH^n.
\end{equation*}
The reverse mapping, $h^{-1}:\BB^n\rightarrow\HH^n$ is defined as follows:
\begin{equation*}
h^{-1}(\bw)=\frac{1}{(1-\norm{\bw}^2)}\begin{bmatrix}
(1+\norm{\bw}^2)\\ 2\bw
\end{bmatrix}, \bw\in\BB^n.
\end{equation*}

Figure~\ref{fig:hyperbolicDiagrams}(c) shows a two-dimensional hyperboloid model $\HH^2$. It can be observed that the \Poincare disk $\BB^2$ is obtained as a stereoscopic projection of $\HH^2$. 

\section{Low-rank parameterization in\\ hyperbolic space}\label{sec:low-rank}
As discussed earlier, hyperbolic embeddings are typically suitable for representing elements of hierarchical structures such as nodes of trees \cite{nickel17a} and complex networks \cite{krioukov10a} to name a few. 
When the task involves closely related hierarchical concepts, additional low-rank structure may also exist among such hyperbolic embeddings. 
In this section, we propose a novel low-rank parameterization for hyperbolic embeddings. 
It should be noted that, unlike the Euclidean embeddings, incorporating a low-rank structure in the hyperbolic framework is non-trivial because of the hyperboloid constraints. 

Let $\bar{\bX}$ be a $(n+1)\times m$ matrix whose columns represent $n$-dimensional  hyperbolic embeddings corresponding to $m$ elements from a given hierarchical structure. 
For notational convenience, we represent $\bar{\bX}$ and its $i$-th column $\bar{\bx}_i$ as follows:
\begin{equation*}
\bar{\bX}=\begin{bmatrix}
\bx_0 \\ \bX
\end{bmatrix} \textup{ and }
\bar{\bx}_i=\begin{bmatrix}
x_{0i}\\ \bx_i
\end{bmatrix}.
\end{equation*}

We propose to approximate $\bar{\bx}_i$ as a low-dimensional hyperbolic embedding such that $\bx_i$ shares a latent low-dimensional subspace with $\bx_j$ (corresponding to $\bar{\bx}_j$), for all $i,j=1,\ldots,m$. 
Mathematically, we propose the following $(r+1)$-rank approximation for $\bar{\bx}_i$:
\begin{equation*}
\bar{\bx}_i
=\begin{bmatrix}
x_{0i}\\ \bx_i
\end{bmatrix}
\approx\begin{bmatrix}
z_{0i}\\ \bU\bz_i
\end{bmatrix}=\hat{\bz}_i\ \ \forall i=1,\ldots,m,
\end{equation*}
where $\hat{\bz}_i\in\HH^n$,  $\bz_i\in\R^r$, $\bU\in\R^{n\times r}$, and $\bU^\top\bU=\bI_r$. We discuss below the consequences of the proposed model. 

Firstly, we obtain $\bar{\bz}_i=\begin{bmatrix}
z_{0i}\\ \bz_i
\end{bmatrix}\in\HH^r$. This is because $\hat{\bz}_i\in\HH^n$ implies 
\begin{equation*}
-z_{0i}^2 + (\bU\bz_i)^\top(\bU\bz_i) = -1.
\end{equation*}
$\bar{\bz}_i\in\HH^r$ follows from the above equality as $\bU^\top\bU=\bI_r$. 
Secondly, the matrix $\bX$ (corresponding to $\bar{\bX}$) is modeled as a low-rank matrix as we approximate $\bX$ as $\bU\bZ$, where $\bZ=[\bz_1,\ldots,\bz_m]$. 
Thirdly, the space complexity of embeddings reduces from $O(nm)$ (for $\bar{\bX}$) to $O(nr+mr)$ (for $\bU,\bZ$ and $\bz_0$). 

We propose to learn the proposed low-rank paramterization of $\bar{\bX}$ by solving the optimization problem:
\begin{equation}\label{eqn:generalOptimization}
\begin{aligned}
& \minop_{\substack{\bU\in\R^{n\times r},\\ \bar{\bz}_i\in\R^{n+1} \forall i}}  & & \underbrace{\sum_{i=1}^m \ell(\bar{\bx_i},\hat{\bz}_i)}_{f(\bU,\bar{\bZ};\bar{\bX})} \\
& \text{subject to} & & \bU^\top\bU=\bI_r, \bar{\bz}_i\in\HH^r, \\
& & &\hat{\bz}_i=\begin{bmatrix} 1 & \bzero_r^\top \\ \bzero_n & \bU \end{bmatrix}\bar{\bz}_i\ \forall i=1,\ldots,m,
\end{aligned}
\end{equation}
where $\ell: \R^{n+1}\times \R^{n+1}\rightarrow [0,\infty)$ is a \textit{loss function} that measures the quality of the proposed approximation. 
Let function $f$ denote the objective function in (\ref{eqn:generalOptimization}), \textit{i.e.}, $f(\bU,\bar{\bZ})=\sum_{i=1}^m \ell(\bar{\bx_i},\hat{\bz}_i)$.

We discuss the following three choices of $f$:
\begin{enumerate}
\item $f(\bU,\bar{\bZ};\bar{\bX})=\|\bX-\bU\bZ\|_F^2$:\newline we penalize the Euclidean distance between $\bX$ and $\bU\bZ$. This is because $\bx_0$ and $\bz_0$ are determined from the hyperboloid constraint given $\bX$ and $\bU\bZ$, respectively. We obtain a closed-form solution of (\ref{eqn:generalOptimization}) with this loss function and the solution involves computing a rank-$r$ singular value decomposition of $\bX$. 
In Section~\ref{sec:experiment}, we denote this approach by the term Method-1. 

\item $f(\bU,\bar{\bZ};\bar{\bX})=\|\bar{\bX}-\hat{\bZ}\|_F^2$:\newline we penalize the Euclidean distance between the (full) hyperbolic embeddings (matrices) $\bar{\bX}$ and $\hat{\bZ}=\begin{bmatrix} 1 & \bzero_r^\top \\ \bzero_n & \bU \end{bmatrix}\bar{\bZ}$. This approach is denoted by the term Method-2 in Section~\ref{sec:experiment}. 

\item $f(\bU,\bar{\bZ};\bar{\bX})=\sum_i \mathrm{arccosh}(-\inner{\bar{\bx}_i,\hat{\bz}_i}_\L)^2$:\newline since the columns of $\bar{\bX}$ and $\hat{\bZ}$ are hyperbolic embeddings, we penalize the hyperbolic distance (\ref{eqn:distance_hyperbolic}) between the corresponding embeddings. We denote it by the term Method-3. 
\end{enumerate}

It should be noted that the problem (\ref{eqn:generalOptimizationf}) is a nonlinear and non-convex optimization problem, but has well-studied structured constraints. In particular, the structured constraints are cast has Riemannian manifolds. In the next section, we propose a Riemannian trust-region algorithm for solving (\ref{eqn:generalOptimization}) with the loss function discussed in options 2) and 3) above.

\section{Optimization}\label{sec:optimization}
It should be noted that the variable $\bU$ in (\ref{eqn:generalOptimization}) belongs to the Stiefel manifold $\Stiefel{r}{n} \coloneqq \{ \bU \in \mathbb{R}^{n\times r} : \bU^\top\bU=\bI_r \}$ \cite{edelman98a} and the variable $\bar{\bz}_i $ belongs to the $r$-dimensional hyperbolic manifold $ \HH^r \coloneqq \{ \bar{\bz}_i \in \mathbb{R}^{r+1} :  -z_{0i}^2 + \bz_i^\top \bz _i= -1,z_{0i} > 0 \}$ for all $i = \{1,\ldots,m\}$. Consequently, the constraint set of the proposed optimization problem (\ref{eqn:generalOptimization}) is a smooth manifold $\mathcal{M} \coloneqq \Stiefel{r}{n} \times \HH^r \times\ldots \times \HH^r $, which is the Cartesian product of the Stiefel and $m$ hyperbolic manifolds of dimension $r$. The problem (\ref{eqn:generalOptimization}), therefore, now boils down to the manifold optimization problem:
\begin{equation}\label{eqn:manifold_optimization}
\begin{array}{lll}
\min\limits_{y \in \mathcal{M}} \quad f(y),
\end{array}
\end{equation}
where $y$ has the representation $y \coloneqq (\bU,\bar{\bz}_1,\ldots, \bar{\bz}_m )$ and $f: \mathcal{M} \rightarrow \mathbb{R} : y \mapsto f(y) = \sum_{i=1}^n \ell(\bar{\bx_i},\hat{\bz}_i)$ is a smooth function.

We tackle the problem (\ref{eqn:manifold_optimization}) in the Riemannian optimization framework that translates it into an unconstrained optimization problem over the nonlinear manifold $\mathcal{M}$, now endowed with a Riemannian geometry \cite{absil08a}. In particular, the Riemannian geometry on manifolds imposes a metric (inner product) structure on $\mathcal{M}$, which in turn allows to generalize notions like the shortest distance between points (on the manifold) or the translation of vectors on manifolds. Following this framework many of the standard nonlinear optimization algorithms in the Euclidean space, e.g., steepest descent and trust-regions, generalize well to Riemannian manifolds in a systematic manner. The Riemannian framework allows to develop computationally efficient algorithms on manifolds~\cite{absil08a}. 

Both the Stiefel and hyperbolic manifolds are Riemannian manifolds, and their geometries have been individually well-studied in the literature \cite{nickel18a,absil08a}. Subsequently, the manifold of interest $\mathcal{M}$ also has a Riemannian structure.

Below we list some of the basic optimization-related notions that are required to solve (\ref{eqn:manifold_optimization}) with the Riemannian \emph{trust-region} algorithm that exploits second-order information. The development of those notions follow the general treatment of manifold optimization discussed in \cite[Chapter~7]{absil08a}. The Stiefel manifold related expressions follow from \cite{absil08a}. The hyperobolic related expressions follow from \cite{nickel18a}. 

\subsection{Metric and tangent space notions}\label{sec:metric}
Optimization on $\mathcal{M}$ is worked out on the tangent space, which is the linearization of $\mathcal{M}$ at a specific point. It is a vector space associated with each element of the manifold.

As $\mathcal{M}$ is a product space, its tangent space is also the product space of the tangent spaces of the Stiefel $\Stiefel{r}{n}$ and hyperbolic $\HH^r$ manifolds. The characterization of the tangent space has the form:
\begin{equation}\label{eqn:tangent_space}
\begin{array}{lll}
T_y \mathcal{M}:= T_\bU \Stiefel{r}{n} \times T_{\bar{\bz}_1}\HH^r \times\ldots \times T_{\bar{\bz}_m} \HH^r  \\
=  \{(\xi_{\bU}, \xi_{\bar{\bz}_i}, \ldots,\xi_{\bar{\bz}_m} ) : {\mathrm {symm}}(\bU^\top \xi_{\bU})= \bzero_r \ \text{ and } \\

\quad  \quad \inner{{\bar{\bz}}_i,\xi_{\bar{\bz}_i}}_\L = 0  \text{ for all }i
 \},
\end{array}
\end{equation}
where $\mathrm{symm}$ extracts the symmetric part of a matrix.

As discussed above, to impose a Riemannian structure on $\mathcal{M}$, a smooth metric (inner product) definition is required at each element of the manifold. A natural choice of the metric $g_y : T_y\mathcal{M} \times T_y\mathcal{M} \rightarrow \mathbb{R} : (\xi_y, \eta_y) \mapsto g_y(\xi_y, \eta_y)$ on $\mathcal{M}$ is the summation of the individual Riemannian metrics on the Stiefel and hyperbolic manifolds. More precisely, we have
\begin{equation}\label{eqn:metric}
\begin{array}{llll}
g_y(\xi_y, \eta_y) \coloneqq \langle \xi_{\bU}, \eta_{\bU} \rangle + \ \sum_{i=1}^m \inner{\xi_{\bar{\bz}_i},\eta_{\bar{\bz}_i}}_\L,
\end{array}
\end{equation}
where $\xi_y = (\xi_{\bU},\xi_{\bar{\bz}_i}, \ldots, \xi_{\bar{\bz}_m} )$, $\eta_y = (\eta_{\bU},\eta_{\bar{\bz}_i}, \ldots, \eta_{\bar{\bz}_m} )$, $\langle \cdot, \cdot \rangle$ is the standard inner product, and $\langle \cdot, \cdot \rangle_\L$ is the Lorentz inner product (\ref{eqn:lorentz_metric}).

 It should be emphasized that the metric in (\ref{eqn:metric}) endows the manifold $\mathcal{M}$ with a Riemannian structure and allows to develop various other notions of optimization in a straightforward manner.

One important ingredient required in optimization is the notion of an orthogonal projection operator $\Pi_y$ from the space $\mathbb{R}^{n\times r}\times \mathbb{R}^{n+1}\ldots \times \mathbb{R}^{n+1}$ to the tangent space $T_y \mathcal{M}$. Exploiting the product and Riemannian structure of $\mathcal{M}$, the projection operator characterization is obtained as the Cartesian product of the individual tangent space projection operator on the Stiefel and hyperbolic manifolds, both of which are well known. Specifically, if $(\zeta_{\bU}, \zeta_{\bar{\bz}_i}, \ldots,\zeta_{\bar{\bz}_m} ) \in \mathbb{R}^{n\times r}\times \mathbb{R}^{n+1}\ldots \times \mathbb{R}^{n+1}$, then its projection onto the tangent space $T_y \mathcal{M}$ is given by
\begin{align}\label{eqn:projection}
\Pi_y(\zeta_{\bU}, \zeta_{\bar{\bz}_i}, \ldots,\zeta_{\bar{\bz}_m} ) \coloneqq &
( \zeta_{\bU} - \bU \mathrm{symm}(\bU ^\top \zeta_{\bU}), \\
&\zeta_{\bar{\bz}_i} + \bar{\bz}_i \inner{{{\bz}}_i,\zeta_{\bar{\bz}_i}}_\L, \nonumber\\
&\ldots,  \nonumber\\
&\zeta_{\bar{\bz}_m} + \bar{\bz}_m \inner{{{\bz}}_m,\zeta_{\bar{\bz}_m}}_\L).\nonumber
\end{align}

\subsection{Retraction}
An optimization algorithm on manifold requires computation of search direction and then following along it. While the computation of the search direction follows from the notions in Section \ref{sec:metric}, in this section we develop the notion of ``moving'' along a search direction on the manifold. This is characterized by the \emph{retraction} operation, which is the generalization of the the exponential map (that follows the geodesic) on the manifold. The retraction operator $R_y$ takes in a tangent vector at $T_y\mathcal{M}$ and outputs an element on the manifold by approximating the geodesic \cite[Definition~4.1.1]{absil08a}. 

Exploiting the product space of $\mathcal{M}$, a natural expression of the retraction operator is obtained by the Cartesian product of the individual retraction operations on the Stiefel and hyperbolic manifolds. If $\xi_y \in T_y \mathcal{M}$, then the retraction operation is given by

\begin{align}\label{eqn:retraction}
 R_y(\xi_y) \coloneqq &
(\mathrm{uf}(\bU + \xi_{\bU}), \\
& {\bar{\bz}}_i\mathrm{cosh}(\| \xi_{{\bar{\bz}}_i}\|_\mathcal{L})+ {\xi_{\bar{\bz}}}_i\mathrm{sinh}(\| \xi_{{\bar{\bz}}_i}\|_\mathcal{L})/\| \xi_{{\bar{\bz}}_i}\|_\mathcal{L}, \nonumber \\
 &   \ldots, \nonumber \\
 &  {\bar{\bz}}_m\mathrm{cosh}(\| \xi_{{\bar{\bz}}_m}\|_\mathcal{L})+ {\xi_{\bar{\bz}}}_m\mathrm{sinh}(\| \xi_{{\bar{\bz}}_m}\|_\mathcal{L})/\| \xi_{{\bar{\bz}}_m}\|_\mathcal{L} ), \nonumber
\end{align}

where $\xi_y = (\xi_{\bU},\xi_{\bar{\bz}_i}, \ldots, \xi_{\bar{\bz}_m} )$, $\| \xi_{{\bar{\bz}}_i}\|_\mathcal{L} = \sqrt{\inner{\xi_{{\bar{\bz}}_i},\xi_{\bar{\bz}_i}}_\L}$ for all $i$, $\mathrm{uf}(\cdot)$ extracts the orthogonal factor of a matrix, \textit{i.e.}, $\mathrm{uf}(\bA) = \bA (\bA^\top \bA)^{-1/2}$ .

\subsection{Riemannian gradient and Hessian computations}\label{sec:gradient_Hessian_computations}
Finally, we require the expressions of the Riemannian gradient and Hessian of $f$ on $\mathcal{M}$. To this end, we first compute the derivatives of $f$ in the Euclidean space. Let $\nabla_y f$ be the first derivative of $f$ and its Euclidean directional derivative along $\xi_y$ is $\mathrm{D} \nabla_y f [\xi_y]$. The expressions of the partial derivatives for the squared Euclidean distance based loss functions mentioned in Section \ref{sec:low-rank} are straightforward to compute. When the loss function is based on the squared hyperbolic distance (\ref{eqn:distance_hyperbolic}), the expressions for $\nabla_y f$ and $\mathrm{D} \nabla_y [\xi_y]$ are discussed in \cite{pennec17a}.  

Once the partial derivatives of $f$ are known, converting them to their Riemannian counterparts on $\mathcal{M}$ follows from the theory of Riemannian optimization \cite[Chapter~3]{absil08a}. The expressions are 
\begin{align}\label{eqn:Riemannian_gradient_Hessian}
\text{Riemannian gradient } &\grad_y f = \Pi_y(\nabla_y f),\ \textup{ and } \\
\text{Riemannian Hessian } &\hess_y f[\xi_y] = \Pi_y ( \mathrm{D}\grad_y f [\xi_y]), \nonumber
\end{align}
where $\Pi_y$ is the orthogonal projection operator (\ref{eqn:projection}).

\subsection{Riemannian trust-region algorithm}
The Riemannian trust-region (TR) algorithm approximates the function $f$ with a \emph{second-order} model at every iteration. The second-order model (which is called the trust-region sub-problem) makes use of the Riemannian gradient and Hessian computations as shown in Section \ref{sec:gradient_Hessian_computations}. The trust-region sub-problem is then solved efficiently (using an iterative quadratic optimization solver, e.g., with the truncated conjugate gradient algorithm) to obtain a candidate search direction. If the candidate search leads to an appreciable decrease in the function $f$, then it is accepted else it is rejected \cite[Chapter~7]{absil08a}. Algorithm \ref{alg:trust_region} summarizes the key steps of the proposed trust-region algorithm for solving (\ref{eqn:manifold_optimization}).

\begin{algorithm}[tb]
   \caption{Riemannian trust-region algorithm for (\ref{eqn:manifold_optimization})}
   \label{alg:trust_region}
   {\center 
   \begin{tabular}{ l  l }
  \multicolumn{2}{l}{{\bfseries Input:} $n$-dimensional hyperbolic embeddings $\bX$ and rank $r$. }   \\
  \multicolumn{2}{l}{ Initialize $y\in\mathcal{M} \coloneqq \Stiefel{r}{n} \times \HH^r \times\ldots \times \HH^r $.}   \\
  \multicolumn{2}{l}{{\bfseries repeat}} \\
  \multicolumn{2}{l}{\ \ \ \ \ \textbf{1:} Compute $\nabla_y f$.}   \vspace{4pt}\\
  
  \multicolumn{2}{l}{\ \ \ \ \ \multirow{2}{500pt}{\textbf{2:} \textbf{Riemannian TR step:} compute a search direction $\xi_y$ which minimizes the trust region sub-problem. It makes use of  $\nabla_y f $ and its directional derivative, and their Riemannian counterparts (\ref{eqn:Riemannian_gradient_Hessian}).} }   \vspace{4pt}\\
 \ \ \ \ \  & \\
   \ \ \ \ \ &  \vspace{1pt}\\
   \multicolumn{2}{l}{\ \ \ \ \ \textbf{3:} Update $y_+ = R_y(\xi_{y})$ (retraction step) from (\ref{eqn:retraction}).} \\
  \multicolumn{2}{l}{{\bfseries until} convergence}   \\
  \multicolumn{2}{l}{{{\bfseries Output:} $y = (\bU, {\bar{\bz}}_1, \ldots, {\bar{\bz}}_m)$ and $\hat{\bZ}=\begin{bmatrix} 1 & \bzero_r^\top \\ \bzero_n & \bU \end{bmatrix}\bar{\bZ}$.}}
\end{tabular}
}
\end{algorithm}

\subsection{Computational complexity}
The manifold-related ingredients cost $O(nr^2 + mr)$. For example, the computation of the Riemannian gradient in (\ref{eqn:Riemannian_gradient_Hessian}) involves only the tangent space projection operation that costs $O(nr^2 + mr)$. Similarly, the retraction operation costs $O(nr^2 + mr + r^3)$.

The computations of $f$ and its derivatives cost $O(nmr)$ (for all the three choices of the loss function in Section \ref{sec:low-rank}). The overall computational cost per iteration of our implementation is, therefore, $O(nmr)$.

\subsection{Numerical implementation}
We use the Matlab toolbox Manopt \cite{boumal14a} to implement Algorithm \ref{alg:trust_region} for (\ref{eqn:manifold_optimization}). Manopt comes with a well-implemented generic Riemannian trust-region solver, which can be used appropriately to solve (\ref{eqn:manifold_optimization}) by providing the necessary optimization-related ingredients mentioned earlier. The Matlab codes are available at \url{https://pratikjawanpuria.com}.

\section{Experiments}\label{sec:experiment}
In this section, we evaluate the performance of the proposed low-rank parameterization of hyperbolic embeddings. In particular, we compare the quality of the low-rank hyperbolic embeddings obtained by minimizing the three different loss functions discussed in Section~\ref{sec:low-rank}.

\subsection*{Experimental setup and evaluation metric} We are provided with the hyperbolic embeddings corresponding to a hierarchical entity such as nodes of a tree or a graph. We also have the ground truth information of the given tree (or graph). Let $(\G,\E)$ represents the ground truth, where $\G$ is the set of nodes and $\E=\{(u,v)\}$ be the set of edges between the nodes ($u,v\in\G$). 
Hyperbolic embeddings can be employed to reconstruct the ground truth since a low hyperbolic distance (\ref{eqn:distance_hyperbolic}) between a pair of nodes implies a high probability of an edge between them. 
However, such a reconstruction may also incorporate errors such as missing out on an edge or adding a non-existent edge. 

We measure the quality of the hyperbolic embeddings as follows: let $u$ and $v$ be a pair of nodes in $\G$ such that $(u,v)\in\E$. Let $\bar{\bu}$ and $\bar{\bv}$ be the  hyperbolic embeddings corresponding to $u$ and $v$, respectively. We compute the hyperbolic distance $d_{\HH}(\bar{\bu},\bar{\bv)}$  (\ref{eqn:distance_hyperbolic}) and rank it among the distance corresponding to all untrue edges from $u$, \textit{i.e.}, $\{d_{\HH}(u,w)|(u,w)\notin\E\}$. We then compute the mean average precision (MAP) of the ranking. The MAP score is a commonly employed metric for evaluating graph embeddings \cite{nickel17a,nickel18a,bordes13a,nickel16a}. Overall, we compare the quality of the proposed low-rank approximation by comparing the MAP score of the original high dimensional embeddings and the low-rank embeddings. 

We obtain the original hyperbolic embeddings from the implementation provided by \cite{nickel17a}. It should be noted that \cite{nickel17a} learns the hyperbolic embeddings from the Poincar{\'e} model and we employ the transformation discussed in Section~\ref{sec:background} to obtain embeddings corresponding to the hyperboloid model. It should be mentioned that though \cite{nickel18a} directly learns hyperbolic embeddings from the hyperboloid model, its implementation is not available. 

\begin{table}\centering
\caption{Mean average precision (MAP) score obtained by the proposed approaches on the mammal dataset.}\label{table:mammal-results}
	\begin{tabular}[t]{rccc}
	\toprule
  	\multicolumn{1}{c}{Rank} & \multicolumn{1}{c}{Method-1} & \multicolumn{1}{c}{Method-2} & \multicolumn{1}{c}{Method-3}\\
	\midrule 
	$5$    & $0.8274$ & $0.8416$ & $0.1516$ \\
	$10$   & $0.9106$ & $0.9143$ & $0.9004$ \\
	$20$   & $0.9388$ & $0.9390$ & $0.9388$ \\ 
	$50$   & $0.9488$ & $0.9486$ & $0.9488$ \\ 
	$100$  & $0.9504$ & $0.9504$ & $0.9504$ \\ 
	$200$  & $0.9502$ & $0.9502$ & $0.9502$\\ 
	$300$  & $0.9501$ & $0.9501$ & $0.9501$\\ 
	\bottomrule
	\end{tabular}
\end{table}

\subsection*{Datasets}
We perform experiments on the mammal and noun subtrees of the WordNet database\cite{miller95}. WordNet is a lexical database and among other things, it also provides relations between pairs of concepts. 

The `mammal' dataset has \textit{mammal} as the root node, with `is-a' (hypernymy) relationship defining the edges. As an example, it has relationships such as `rodent' \textit{is-a} `mammal', `squirrel' \textit{is-a} `rodent', \textit{etc}. Hence, there exists an edge from the `mammal' node to `rodent' node and from `rodent' node to `squirrel' node. The WordNet mammal subtree consists of $|\G|=1\,180$ nodes and $|\E|=6\,540$ edges.  A part of this subtree is displayed in Figure~\ref{fig:hyperbolicDiagrams}(a). 

Similarly, the `noun' dataset is also a subtree of WordNet database. Examples in this subtree include `photograph' \textit{is-a} `object', 'bronchitis' \textit{is-a} `disease', `disease' \textit{is-a} `entity', \textit{etc}. 
It consists of $|\G|=82\,115$ nodes and $|\E|=743\,086$ edges. 

\subsection*{Results}
We compare the performance of the proposed low-rank approximation of hyperbolic embeddings with the three loss functions discussed in Section~\ref{sec:low-rank}. 
Table~\ref{table:mammal-results} reports the results on the mammal dataset with different values of rank $r=\{5,10,20,50,100,200,300\}$. The original $300$-dimensional hyperbolic embeddings for the mammal subtree achieve a MAP score of $0.9501$. We observe that all the three methods are able to obtain MAP scores very close to the original embeddings with rank $r\geq50$. In addition, Method-1 and Method-2 perform well even in very low-rank setting ($r=5$). This hints that penalizing with the Euclidean distance may be more suitable than compared to the hyperbolic distance (\ref{eqn:distance_hyperbolic}) for approximating hyperbolic embeddings when the given rank is very small. 

The results on the noun dataset are reported in Table~\ref{table:noun-results}. This dataset is challenging because of its scale and relatively low reconstruction performance of the original hyperbolic embeddings.  The original $100$-dimensional hyperbolic embeddings for the noun subtree achieve a MAP score of $0.8070$. We observe that at rank $r=20$ our methods are able to get within $90\%$ of the performance obtained by the original embeddings. 
\begin{table}\centering
{
\caption{Mean average precision (MAP) score obtained by the proposed approaches on the noun dataset.}\label{table:noun-results}
	\begin{tabular}[t]{rccc}
	\toprule
  	\multicolumn{1}{c}{Rank} & \multicolumn{1}{c}{Method-1} & \multicolumn{1}{c}{Method-2} & \multicolumn{1}{c}{Method-3}\\
	\midrule 
	$5$    & $0.5343$ & $0.5422$ & $0.5343$ \\
	$10$   & $0.6742$ & $0.6796$ & $0.6742$ \\
	$20$   & $0.7425$ & $0.7449$ & $0.7425$ \\ 
	$50$   & $0.7887$ & $0.7891$ & $0.7887$ \\ 
	$100$  & $0.8070$ & $0.8070$ & $0.8070$ \\ 
	\bottomrule
	\end{tabular}
}
\end{table}

\section{Conclusion and Future work}\label{sec:conclusion}
Recently, hyperbolic embeddings have gained popularity in many machine learning applications because of their ability to model complex networks. In this paper, we have looked at scenarios where hyperbolic embeddings are potentially high dimensional and how to compress them using a low-rank factorization model. While low-rank decomposition of Euclidean embeddings are well-known, that of hyperbolic embeddings has not been well-studied. To this end, we have proposed a systematic approach to compute low-rank approximations of hyperbolic embeddings. Our approach allows to decompose a high dimensional hyperbolic embedding ($\bar{\bx}$) into a product of low-dimensional subspace ($\bU$) and a smaller dimensional hyperbolic embedding ($\bar{\bz}$).

We modeled the learning problem as an optimization problem on manifolds. Various optimization-related notions were presented to implement a Riemannian trust-region algorithm. Our experiments showed the benefit of the proposed low-rank approximations on real-world datasets.

As a future research direction, we would like to explore how low-rank hyperbolic embeddings are useful in downstream applications. Another research direction could be on developing methods to compute a ``good'' rank of hyperbolic embeddings.

\bibliographystyle{unsrt}
\bibliography{JMM_arXiv}

\end{document}